\pdfoutput=1

\documentclass[11pt]{article}

\usepackage{acl}

\usepackage{times}
\usepackage{latexsym}

\usepackage[T1]{fontenc}

\usepackage[utf8]{inputenc}

\usepackage{microtype}

\usepackage{inconsolata}

\usepackage{amsmath, amsfonts, amssymb, amsbsy, mathtools, nccmath, bm}

\usepackage{graphicx}
\usepackage{subcaption}
\usepackage{afterpage}

\usepackage{multirow}
\usepackage{booktabs}
\usepackage{dblfloatfix}

\usepackage{dsfont}

\makeatletter
  \newcommand\scripty{\@setfontsize\scripty{6pt}{7}}
\makeatother

\usepackage{hyperref}
\definecolor{darkblue}{rgb}{0.0, 0.0, 0.45}
\hypersetup{colorlinks=true,linkcolor=darkblue,citecolor=darkblue,filecolor=darkblue,urlcolor=darkblue,pagebackref=true,breaklinks=true,bookmarks=true}

\usepackage[multiple]{footmisc}
\makeatletter
\renewcommand\@makefntext[1]%
    {\makebox[4pt][r]{\textsuperscript{\@thefnmark}\,}#1}
\makeatother

\usepackage{xcolor}

\usepackage{thanks-nostar}

\title{Detecting concrete visual tokens for Multimodal Machine Translation}

\author{Braeden Bowen$^{\text{1}}$, Vipin Vijayan$^{\text{1}}$, Scott Grigsby$^{\text{1}}$, Timothy Anderson$^{\text{2}}$,
Jeremy Gwinnup$^{\text{2}}$\\
$^{\text{1}}$PAR Government Systems Corporation, $^{\text{2}}$Air Force Research Laboratory\\
\texttt{\small \{braeden\_bowen, vipin\_vijayan, scott\_grigsby\}@partech.com},\\\texttt{\small \{timothy.anderson.20, jeremy.gwinnup.1\}@us.af.mil}}

\begin{document}
\maketitle
\begin{abstract}
The challenge of visual grounding and masking in multimodal machine translation (MMT) systems has encouraged varying approaches to the detection and selection of visually-grounded text tokens for masking. We introduce new methods for detection of visually and contextually relevant (concrete) tokens from source sentences, including detection with natural language processing (NLP), detection with object detection, and a joint detection-verification technique. We also introduce new methods for selection of detected tokens, including shortest $n$ tokens, longest $n$ tokens, and all detected concrete tokens. We utilize the GRAM MMT architecture to train models against synthetically collated multimodal datasets of source images with masked sentences, showing performance improvements and improved usage of visual context during translation tasks over the baseline model.

\end{abstract}

\section{Introduction}
\label{sec:intro}
The challenge of multimodal machine translation (MMT) is to design a system that automatically translates text from one language to another while utilizing other modalities (e.g., image, video, audio) as inputs to assist in translation \cite{caglayan-multi-2016}. %

Prior work has shown that translation ambiguities and missing textual information can be supplied by contextually-relevant images, aiding in multilingual translation \cite{lala-specia-2018-multimodal, caglayan-etal-2019-probing, wu-good-2021}. For example, the noun ``bank'' is ambiguous and contextually dependent in English (``financial institution'' or ``river edge'') but unambiguous in French (``\textit{banque}'' or ``\textit{rive}'') \citep{futeral-etal-2023-commute}. The hypothesis for MMT research is that these translation ambiguities can be resolved with the inclusion of image context.

In practice, not every sentence has semantic ambiguities, missing information, or relevant visual context; it is therefore beneficial to ensure that ambiguous text is visually and contextually relevant to an associated image \citep{zhou-visual-2018}.

To enforce reliance on image context for translation tasks, some MMT models mask tokens from text inputs \citep{caglayan-etal-2019-probing, sato-etal-2023-choosing}. While most early masking iterations randomly selected tokens for masking, more recent efforts have sought to mask tokens based on contextual relevance to a given image \citep{tan-2020-vokenization}, increasing the usefulness of the image in resolving ambiguity. Still, those methods tend to ignore deterministic selection of relevant tokens, opting to randomly select from a pool of viable tokens.

While these approaches have displayed performance improvements over text-only and random masking models, these methods generally do not take into account the relevance of a masked token. Therefore, we hypothesize that more intentional selection and masking of \textbf{concrete} (i.e., visually and contextually relevant) text tokens, will improve visual grounding and increase model usage of multimodal context. 

In order to select visually and contextually relevant tokens, we explore a combination of natural language processing (NLP) techniques and object detection models and examine deterministic methods for selection of tokens from the available detections. 

Using these techniques, we collate multimodal datasets based on the Multi30k dataset \citep{elliott-2016-multi30k}; the resulting datasets are triplets of source sentences with masked concrete tokens, unmasked target sentences, and associated images.

When masking concrete text tokens from source sentences, we find improvements in both usage of visual information in translation and in performance on evaluation challenges, including CoMMuTE scores of up to $0.67$ and BLEU scores of up to $46.2$.

\begin{figure*}
    \centering
    \includegraphics[width=1.0\textwidth, trim={0cm 2cm 0cm 0cm}, clip=true]{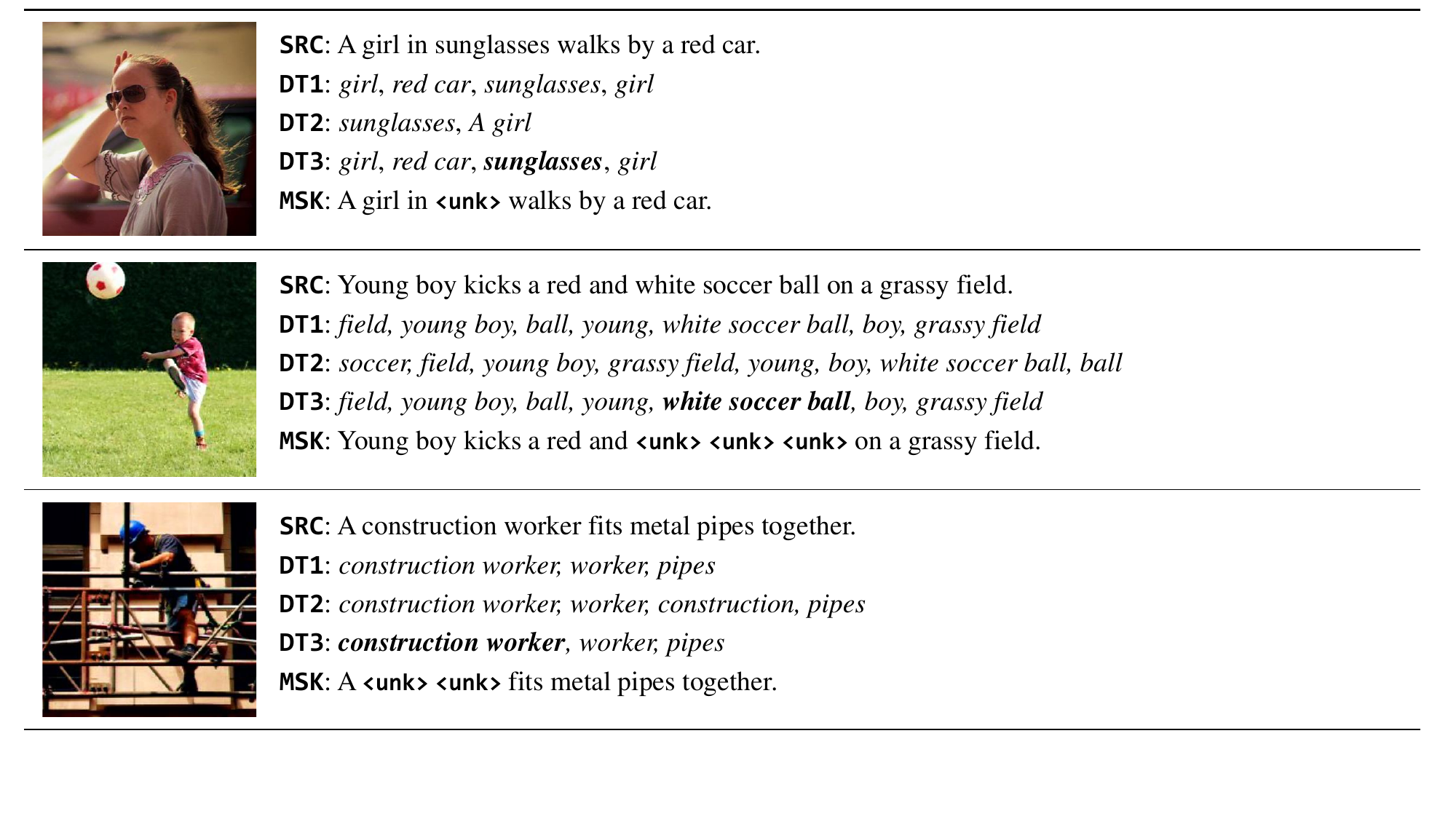}
    \caption{Multi30k source pairs (image, \textbf{SRC}) with results from each detection technique (\textbf{DT}) and an example masked source text (\textbf{MSK}). \textbf{DT1} represents the \textit{NLTK} technique; \textbf{DT2} represents the \textit{MDETR Detection} technique; \textbf{DT3} represents the \textit{Joint Detection} technique. The masked sentence \textbf{MSK} represents a possible masked sentence based on the bold token in the \textbf{DT3} detections.}
    \label{tab:examples}
\end{figure*}

\section{Related Works}
\label{sec:related}

\subsection{Masking for Visual Grounding}
In a text-only modality, \citet{devlin-etal-2019-bert} randomly masked text tokens during pre-training of a bidirectional transformer encoder-decoder and found performance improvements against other text-only models. %

\citet{zhou-visual-2018} utilized jointly-encoded unmasked text and image embeddings to visually ground entire source sentences to images. Using a visual-text attention mechanism on the embeddings, they extracted words that shared semantic context with the images.

\citet{ive-distilling-2019} combined these approaches, randomly \textit{and} manually masking ambiguous and gender-neutral words from source texts to force their MMT model to utilize visual information on evaluation tasks. This work showed that the model was able to use image context to recover from missing, inaccurate, or ambiguous textual context.

\citet{caglayan-etal-2019-probing} used image descriptions from the Flicker30k-Entities dataset \cite{plummer-entities-2015} to dynamically mask \textit{visually depictable entities} and color descriptors from source sentences, but noted a degradation in performance on the Multi30k test sets \cite{elliott-2016-multi30k}. In contrast, \citet{wang-efficient-2021} found that masking \textit{irrelevant} objects improved performance on MMT evaluation tasks, suggesting that state-of-the-art MMT models are ineffectively utilizing visual information.

A meta-analysis by \citet{wu-good-2021} found that many reported improvements in MMT performance are the result of regularization effects, not model interpolation of multimodal features; similarly, \citet{zhuang-visual-2023} found that while visual grounding can improve performance in word learning, these improvements are only marginal. However, they also found that training sets with less textual information and fewer direct co-occurrences of visual words more effectively utilize visual information, suggesting that the relationship between text and image context is still viable.

\subsection{Token Selection for Visual Grounding}
In practice, many sentences have more than one visually grounded token; in these cases, available tokens must be dynamically selected for masking. The standard method is to randomly select viable tokens \cite{devlin-etal-2019-bert}; however, recent work in masked language modeling (MLM) has shown that informed selection of masked tokens may improve performance \citep{sato-etal-2023-choosing}.

Other work has given consideration to the length of source segments in text masking \citep{xiao2023amom} and to the number of tokens selected \citep{tacl_a_00300}, but little work has been done to select tokens deterministically (e.g., by token length).

\section{Approach}
\label{sec:approach}

We perform improved visual grounding by detecting concrete tokens in source sentences. We explore three detection techniques to identify concrete text tokens (Section \ref{sec:detection}) and four selection techniques to appropriately select the identified concrete text tokens (Section \ref{sec:selection}). We then collate permutations of synthetic MMT datasets by masking the selected concrete tokens from source sentences and aligning each sentence with its original dataset image pair. We then train an MMT model (Section \ref{sec:gram}) on these datasets, expanding on work by \citet{vijayan-multimodal-2024} and \citet{caglayan-etal-2019-probing}. %

\subsection{Detection of Concrete Tokens}
\label{sec:detection}
As \citet{caglayan-etal-2019-probing} found, masking visually relevant objects from a source text can force the model to utilize image context to fill in the artificially-created gap in lexical/semantic understanding. We hypothesize that for a given text-image pair, the masking of text tokens that are directly relevant to the image (i.e., ``concrete'' tokens), will improve visual grounding, increasing model correlation of image inputs during downstream translation tasks.

We present three techniques for detection of concrete tokens: NLP with NLTK (Section \ref{sec:nltk}), object detection with MDETR (Section \ref{sec:mdetr}), and joint NLTK/MDETR detection and grounding (Section \ref{sec:joint}). While techniques one and two respectively use text and image context, method three uses contextual information from both modalities to make decisions about which text tokens are concrete. 

\subsubsection{Detection with NLTK}
\label{sec:nltk}
The first concrete token detection approach is to parse sentences for nouns and noun phrases that are likely to represent visual context. By masking tokens that are critical to comprehension and translation of the text, we can encourage the model to learn with visual context.

\begin{figure}[htbp]
    \centering
    \includegraphics[width=1.0\linewidth, trim={4.5cm 16cm 7cm 5cm}, clip=true]{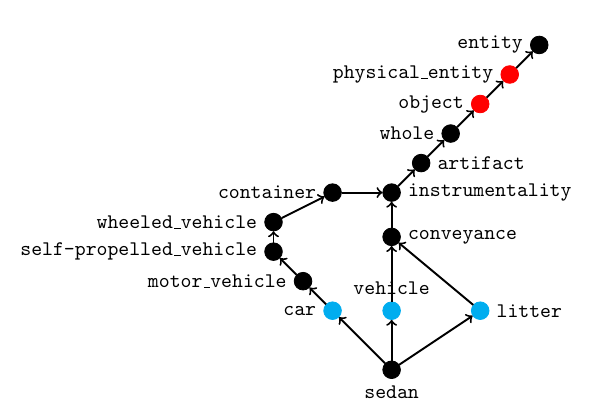}
    \caption{An example hypernym graph. The original token, \texttt{sedan}, its three synset entries (labeled in blue), and its associated concrete hypernyms (labeled in red).}
    \label{fig:dag}
\end{figure}

The Natural Language Toolkit (NLTK) \cite{loper-nltk-2002} includes the WordNet corpus \cite{miller_wordnet_1995}, an English-language lexical database that provides structured relationships between cognitive synonyms (``\textit{synsets}'') for nouns, verbs, adjectives, and adverbs. Specifically, WordNet defines a directed acyclic graph (DAG) for each of these parts of speech (POS), containing synonyms, troponyms, antonyms, and meronyms (Figure \ref{fig:dag}). Critically, these relational graphs establish affiliations between synsets and hypernyms: that is, English words, their definitions, and their related parent categories.

Starting with specific synonyms and troponyms (e.g., ``\texttt{sedan}'', ``\texttt{hatchback}'', ``\texttt{SUV}'') and traversing the DAG upwards, WordNet collapses definitions and synsets into their associated hypernym classes (e.g., ``\texttt{car}'', ``\texttt{vehicle}'') until it reaches a root hypernym (e.g., ``\texttt{physical\_entity}'', ``\texttt{entity}''). Using recursive graph traversal, we can select any node in the DAG and parse its hypernyms upward until we reach either a root hypernym or a parent hypernym on which we can base an estimate of the root hypernym (e.g., ``\texttt{object}'' generally maps to ``\texttt{physical\_entity}'').

Given that there exists only a small cluster of root and high-level parent hypernyms for nouns in WordNet, we can classify the hypernym DAG of any noun or noun phrase as ``\textbf{concrete}'' or ``\textbf{abstract}'' based on these high-level hypernyms (Table \ref{tab:hyper}).

While this method provides a simple concrete/abstract classifier for text tokens, it introduces additional complications. Although most DAG nodes have multiple child hyponyms (e.g., ``\texttt{car}'' may have ``\texttt{sedan}'' and ``\texttt{hatchback}''), some have multiple cognitive synonyms, as English words often have multiple equally likely definitions. For a given node, each of its ``definitions'' will appear as an entry into its synset; for example, the English noun ``link'' has nine values in its WordNet synset, ranging from ``\textit{URL}'' to ``\textit{channel for communication}'' to ``\textit{element of a chain}.'' These varied definitions may branch to different root hypernyms, impacting the classification based on which definition is chosen (Table \ref{tab:hyper}).

To compensate, we consider each entry in a word's synset and extract a ratio of concrete/abstract definitions, which more comprehensively projects a token's likelihood of being concrete. We perform recursive graph traversal for each entry and retain the percent of concrete entries as a ``concreteness score.'' To then classify the original word as abstract or concrete, we establish a threshold of $33\%$ likelihood and only accept words above that concreteness score.
 
\subsubsection{Detection with MDETR}
\label{sec:mdetr}
While the NLTK approach can quickly and efficiently select concrete tokens from a sentence, it incorrectly assumes that every concrete token in the sentence is relevant to its associated image. Contextually linking an irrelevant concrete token to a given image could negatively impact model performance, especially if the token has high commonality in a dataset. As a second approach to concrete token detection, we utilize an object detection model to select concrete tokens. Rather than relying solely on the text processing for detection, we inspect the image itself for object classes relevant to the text.

\begin{figure}[th]
    \centering
    \includegraphics[width=1.0\linewidth, trim={1.125cm 7cm 7cm 9cm}, clip=true]{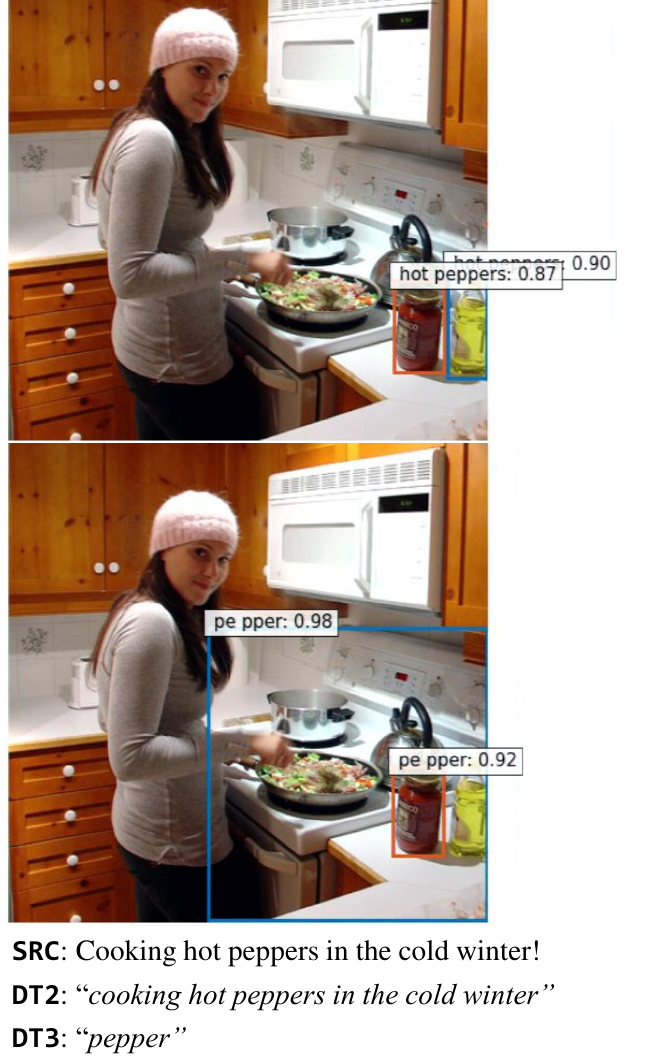}
    \caption{Multi30k source pair (image, \textbf{SRC}) with results from the MDETR (\textbf{DT2}, top image) and Joint (\textbf{DT3}, bottom image) detection techniques. MDETR query strings, bounding boxes, and confidence scores are shown. In this example, supplying the entire source sentence as text input to the MDETR object detection model incorrectly identifies the peppers being cooked, while querying only the word ``\textit{pepper}'' more closely identifies the region containing the query.}
    \label{fig:mdetrjoint}
\end{figure}

For this approach, we use MDETR \cite{kamath-2021-mdetr}, an end-to-end object detection model. Rather than relying exclusively on pre-defined object classes, MDETR uses NLP techniques alongside a pre-trained detection model \cite{carion2020endtoend} to perform object detection and image classification based on the input tokens. Given a text-image pair (Figure \ref{fig:mdetrjoint}), the model assigns each text token an object classification and assigns it confidence score and bounding box within the image. To maximize the number of detectable tokens, we pass an entire Multi30k sentence into the MDETR model and filter out detections with low confidence scores, retaining only the tokens with a high confidence of correlation to the image. While \citet{kamath-2021-mdetr} filter all outputs with confidence less than $0.7$, we filter at $0.85$; after analyzing performance at threshold increments between $0.5$ and $0.95$, we found that this threshold ensured the most balanced confidence in image objects.

\subsubsection{Detection with Joint Visual Grounding}
\label{sec:joint}
While the MDETR technique is less likely than the NLTK technique to improperly select text tokens as visually-grounded, the pre-trained MDETR model will always attempt to match text tokens with a bounding box in the image, often resulting in outputs with high confidence but incorrect alignment. In practice, providing extended textual context (i.e., entire captions or sentences) further exacerbates this problem (Figure \ref{fig:mdetrjoint}).

Therefore, we are left with two techniques with contrasting weaknesses: NLTK ignores image context, and MDETR misinterprets textual context. To mitigate these issues, we present a conjoined detection technique that ``verifies'' the presence of NLTK-detected concrete tokens within an image using MDETR, ensuring that concrete tokens are visually grounded in the image.

Like the MDETR technique, the joint technique parses text-image pairs (unlike the NLTK technique, which is image-agnostic). The source sentence is first processed by the NLTK technique, which returns the noun and noun phrase tokens that met or surpassed the concrete threshold. Each of those tokens is paired with a copy of the source image and passed into the MDETR technique, which performs object detection and filters out all tokens whose resulting confidence is below the confidence threshold. This simultaneously reduces the probability of incorrect alignment by the object detection model and ensures that text tokens are visually grounded, resulting in a set of linguistically concrete and visually-grounded text tokens with high probability of relevance to the source image. Masking these explicitly-relevant tokens will force model reliance on image context.

\begin{figure}[h]
    \centering
    \includegraphics[width=\linewidth, trim={0cm 0cm 0cm 0cm}, clip=true]{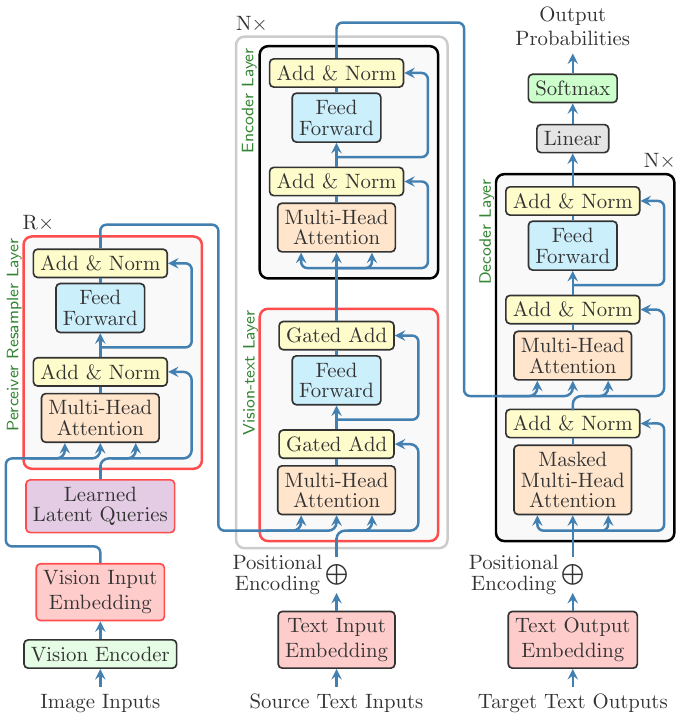}
    \caption{GRAM model architecture from \citet{vijayan-multimodal-2024}.}
    \label{fig:gram}
\end{figure}

\subsection{Synthetic Dataset Collation}
\label{sec:dataset}
Because most current work in MMT focuses on the Multi30k dataset \citep{elliott-2016-multi30k}, an image-caption dataset consisting of 30,014 images with English sentences and corresponding multilingual translations, we collate synthetic datasets of masked sentence-image pairs from Multi30k.

We use each detection technique (Section \ref{sec:detection}) to detect concrete tokens and align them to their original dataset image. From these masked sentence-image pairs, we collate a series of MMT datasets in which a maximum of two concrete tokens are masked from each sentence and associated with the relevant image from the original dataset, resulting in training and validation sets that are at most twice as large as the original Multi30k sets. 

\subsubsection{Token Selection Techniques}
\label{sec:selection}
During the dataset collation process, a single sentence may have more than $n$=$2$ available concrete tokens; in this case, additional consideration must be given to which tokens are selected for inclusion in the dataset. The standard method has generally been to randomly select from the available tokens \cite{devlin-etal-2019-bert}, but recent work in masked language modeling (MLM) has shown that more informed selection of masked tokens may actually improve performance \citep{sato-etal-2023-choosing}.

To examine this, we implement two deterministic token selection techniques, selecting the $n$ \textbf{longest} and \textbf{shortest} tokens (by number of characters) respectively for each sentence. We compare these techniques to a \textbf{random} selection of $n$ tokens and an \textbf{unrestricted} selection which ignores the $n$=$2$ normalization and accepts all available concrete tokens.

\subsection{GRAM Model}
\label{sec:gram}
As the basis for our multimodal translation architecture, we utilize the GRAM architecture \cite{vijayan-multimodal-2024}. GRAM modifies the FAIR WMT19 \cite{ng-fair-2019} text-only model, an encoder/decoder-based transformer architecture \cite{vaswani-2017-attention}, by adding additional multimodal components (Figure \ref{fig:gram}) to create an MMT model.

To process text input, GRAM uses the same byte-pair encoding (BPE) and vocabulary dictionary as the FAIR WMT19 model \cite{ng-fair-2019}. Masked sentences are BPE-encoded and fed as standard text inputs to the MMT model. We mask by replacing each token with an \texttt{<unk>} token, as that token is the closest to a mask token available in the FAIR WMT19 model \cite{ng-fair-2019}. Our method expands on prior work by \citet{tang-etal-multimodal-2022} and \citet{wu-good-2021} while increasing the requirements for a token to be visually grounded to an image.

To process image input, the GRAM model uses CLIP, a pre-trained text-only translation model alongside a pre-trained vision encoder, a perceiver resampler, and vision-text cross-attention layers \citep{radford-clip-2021}. While the original GRAM paper utilizes the \texttt{ViT-L/14@336px} CLIP model, we noted better results within our evaluation framework when using the \texttt{RN50x4} CLIP model; we present those results below (Section \ref{sec:results}). This vision encoder converts input images into image embeddings, enabling the perceiver resampler to convert those embeddings into a fixed number of vision tokens. Vision tokens and corresponding text embeddings are interleaved into vision-text cross-attention layers within the transformer encoder, creating mappings from both the text and the image embeddings onto a sequence of joint representations. Finally, the transformer decoder ingests this sequence and outputs probabilities for the next output text token in the target sequence.

The number of parameters in the original text-only Transformer is 269,746,176; the number of parameters in the RN50x4 CLIP vision encoder is 101,520,396, for a total of 371,266,572 parameters in our GRAM model. Additionally, our GRAM perceiver resampler contains 87,137,080 parameters.

\begin{table*}[ht]
    \centering
    
\begin{tabular}{llcccc}\toprule
    \multicolumn{1}{c}{\textbf{Detection}}&\multicolumn{1}{c}{\textbf{Selection}}& \multicolumn{4}{c}{\textbf{Score}} \\
    \cmidrule(lr){3-6}
    \multicolumn{2}{c}{ } & CoMMuTE & \multicolumn{3}{c}{Multi30k BLEU4 (en-de)}\\
    \cmidrule(lr){4-6}
    \multicolumn{3}{c}{} & 2016 & 2017 & COCO \\\midrule

    \multicolumn{2}{c}{\citet{futeral-etal-2023-commute}}&\textit{0.59}&\textit{43.3}&\textit{38.3}&\textit{35.7}\\ %
    \multicolumn{2}{c}{\citet{anon-gram-2024}}&\textit{0.61}&\textit{\textbf{46.5}}&\textit{\textbf{43.6}}&\textit{39.1}\\\midrule %
    
    Unmasked&&0.5&45.0&42.0&38.2\\\midrule
    
    NLTK&Unrestricted&0.55&45.7&41.9&\underline{\textbf{39.2}}\\
    NLTK&Restricted-Longest&0.62&46.0&\underline{42.5}&37.8\\
    NLTK&Restricted-Shortest&0.63&46.0&42.0&37.9\\
    NLTK&Restricted-Random&\textbf{\underline{0.67}}&\underline{46.2}&41.4&37.8\\\midrule
    
    MDETR&Unrestricted&0.56&\underline{46.0}&\underline{42.4}&\underline{38.4}\\
    MDETR&Restricted-Longest&0.63&45.7&41.7&38.0\\
    MDETR&Restricted-Shortest&0.63&45.0&41.2&36.9\\
    MDETR&Restricted-Random&0.63&45.6&42.2&37.6\\\midrule
        
    Joint&Unrestricted&0.52&45.5&42.4&\underline{38.9}\\
    Joint&Restricted-Longest&\underline{0.63}&\underline{45.8}&\underline{42.6}&38.8\\
    Joint&Restricted-Shortest&0.61&45.4&42.0&37.9\\
    Joint&Restricted-Random&0.61&45.5&42.4&37.5\\\bottomrule
\end{tabular}
    \caption{Selected performance results of our model against the CoMMuTE and Multi30k test sets. The best result by column is indicated in \textbf{bold}; the best result for each detection technique is \underline{underlined}. Results as reported by GRAM \cite{vijayan-multimodal-2024} and VGAMT \cite{futeral-etal-2023-commute} are included for reference.}
    \label{tab:results}
\end{table*}

\section{Results and Discussion}
\label{sec:eval}

\subsection{Experimental Framework}
We train the GRAM models on unique permutations of synthetically collated datasets representing each combination of detection (\textbf{NLTK}, \textbf{MDETR}, \textbf{Joint}) (Section \ref{sec:detection}) and selection (\textbf{unrestricted}, \textbf{restricted-long}, \textbf{restricted-short}, \textbf{restricted-random}) (Section \ref{sec:selection}) techniques. We compare the resulting trained versions to the GRAM model trained on a unmasked dataset of original sentences.

Most current work in MMT focuses on the Multi30k dataset; because of its prevalence in other MMT works, we utilize the Multi30k dataset for collation of our training datasets. We then evaluate the GRAM models on the Multi30k 2016, 2017, and COCO test sets using BLEU4 scores. 

We also evaluate the GRAM model with an additional metric, Contrastive Multilingual Multimodal Translation Evaluation (CoMMuTE). \citet{futeral-etal-2023-commute} proposed the CoMMuTE dataset to evaluate both performance on translation tasks and usage of visual information by MMT models. In the ensemble CoMMuTE evaluation, the model is given two images, a lexically or semantically ambiguous English sentence, and a target language translation that resolves the ambiguity according to one of the two images. The task involves determining which of the two images the sentence pairs best match. The evaluation is made using the perplexity of the model output, and the resulting CoMMuTE score is calculated using the model's determination of accuracy across 100 text-image pairs.

\begin{table}[ht]
    \centering
    \begin{tabular}{lcc}\toprule
    \textbf{Detection}&\textbf{Concrete \%}&\textbf{Unique Detections}\\\midrule
     
     \textbf{NLTK}&99.51\%&5,393\\
     \textbf{MDETR}&99.92\%&6,674\\
     \textbf{Joint}&99.49\%&4,761\\\bottomrule
\end{tabular}
    \caption{Unique concrete token detections and percent of Multi30k sentences with detected tokens by detection technique.}
    \label{tab:det_data}
\end{table}

\subsection{Results}
\label{sec:results}
We review the performance of the model variants trained using the synthetic Multi30k datasets (Section \ref{sec:dataset}) on the above evaluation metrics. We train $13$ variants, consisting of one unmasked baseline and $12$ models representing each combination of detection (Section \ref{sec:detection}) and selection (Section \ref{sec:selection}) techniques. 

\subsection{Detection Results}
\label{q1}
We introduced three distinct methods for detection of concrete text tokens: the NLTK technique (Section \ref{sec:nltk}), which parses nouns and noun phrases from sentences, the MDETR technique (Section \ref{sec:mdetr}), which inputs sentences as queries to an object detection model, and the Joint technique (Section \ref{sec:joint}). Each technique generates the same output structure: multimodal datasets of sentences masked concrete tokens and matching images. We hypothesize that masking concrete tokens with these techniques will improve performance on evaluation metrics. We further hypothesize that the Joint technique will be more selective with its detections than its component NLTK and MDETR techniques, and will thus utilize image context more efficiently and critically.

We found that all three techniques consistently extracted relevant tokens from the text: each technique extracted concrete tokens from over $99\%$ of Multi30k sentences (Table \ref{tab:det_data}). The MDETR detection technique was the most successful, extracting $23.8\%$ and $40.2\%$ more unique concrete tokens than the NLTK and Joint techniques, respectively. This resulted in the MDETR technique masking the highest concentration of original Multi30k sentences ($114$ and $120$ sentences more than NLTK and Joint, respectively).

Increased rates of detection did not correlate with better performance, though. All tested models outperformed the unmasked (baseline) dataset in CoMMuTE and BLEU scores, but in contrast to our hypothesis the NLTK technique outperformed both the MDETR and Joint techniques both in CoMMuTE and BLEU score (Table \ref{tab:results}, \ref{tab:results}). The Joint technique, which we hypothesized would improve on its component techniques, consistently underperformed against the others. This is especially true in the \texttt{Joint Unrestricted} model, which only improved its CoMMuTE score by $0.02$ and its BLEU score $0.5$ over the baseline. We suggest that the Joint technique was actually hindered by its strict selection process, leading to a much smaller pool of objects to mask from. Conversely, the MDETR technique tended to over-select longer, rarely-used, or irrelevant tokens (Figure \ref{fig:mdetrjoint}), contributing to the larger masking percentages but the lower overall performance. The success of the NLTK technique over the others was its ``middle ground'' approach, classifying concrete tokens more liberally than the Joint technique but more consistently than the MDETR technique.

$23\%$ of tested models underperformed the original GRAM model \cite{vijayan-multimodal-2024} on CoMMuTE metrics, $15.4\%$ performed identically, and the remaining $53.8\%$ outperformed. All tested models underperformed the original GRAM model in Multi30k 2016/2017 BLEU metrics. One model (\texttt{NLTK Unrestricted}) outperformed the original GRAM model in the Multi30k COCO metric, but the improvement is well within a margin for normalization effects. We suggest that the performance disparity between models in these Multi30k BLEU metrics is due to dataset size: the original GRAM model was pre-trained trained on the Conceptual Captions dataset \cite{sharma-etal-2018-conceptual} of $2,878,999$ text-image pairs, resulting in synthetic datasets nearly 100 times larger than those used here. Despite this, the majority of models outperformed GRAM in CoMMuTE metrics, achieving scores of up to $0.67$.

In general, we also note an inverse relationship between CoMMuTE and BLEU performance: that is, when CoMMuTE scores increase, BLEU scores tend to decrease. For example, the \texttt{MDETR Unrestricted} model notched the highest average BLEU score across all three Multi30k metrics, but had the second-lowest CoMMuTE score. 

\subsection{Selection Results}
\label{q2}

Critical to the synthetic dataset collation system is the process of selecting concrete tokens for masking. Prior efforts have generally selected tokens at random \cite{ive-distilling-2019}; we introduced three additional techniques (Section \ref{sec:selection}), longest-token selection, shortest-token selection, and unrestricted selection, and test each against a baseline of randomly-selected concrete tokens. We hypothesize that the presented token selection techniques will outperform the baseline of random selection; specifically, we hypothesize that longest-token and unrestricted selection will encourage additional usage of visual context and thus improve CoMMuTE score, and that shortest-token selection will minimize the number of token predictions required by the model (Section \ref{sec:gram}) and thus improve BLEU score.

We found that while all tested selection techniques (Section \ref{sec:selection}) outperformed the unmasked baseline, comparative performance between techniques are less conclusive. When paired with the NLTK detection technique, the random selection technique outperformed the others in CoMMuTE and Multi30k 2016 BLEU scores. When paired with the MDETR metric, none of the restricted selection techniques had any impact on CoMMuTE score. When paired with the Joint detection technique, the longest-token selection technique improved CoMMuTE and Multi30k 2016/2017 BLEU scores.

Contrary to our hypothesis, the deterministic token selection techniques did not consistently outperform the random selection technique. The most consistent results were with the unrestricted selection technique, which significantly degraded CoMMuTE performance but tended to improve BLEU performance (especially in the COCO BLEU metric, where it outperformed all other tested models). Shortest-token selection also tended to follow these patterns of performance degradation, but not as substantially: its NLTK and Joint detection variants performed identically on the Multi30k 2017 and COCO BLEU metrics and performed near the bottom of results for the CoMMuTE and 2016 BLEU metrics across all three detection techniques.

Each of these findings runs counter to our hypotheses in this area, suggesting that token selection at this scale has less impact on model performance than we expected; in fact, random or pseudo-random token selection of the identified concrete tokens may actually improve performance over deterministic methods.

\subsection{Future Work}
\label{sec:future}
Given the high percentage of visually-grounded tokens in the Multi30k training set, future work should consider the techniques against both larger MMT datasets and MMT datasets with lower concentrations of visually-grounded tokens (e.g., Conceptual Captions). Similarly, future work should consider synthetically collated datasets that combine elements of multiple multimodal datasets (e.g., images from Conceptual Captions, sentences from Multi30k), including synthetic datasets created from text-only datasets.

Additionally, future work should compare baseline scores for tokens selected completely at random to more accurately gauge the efficacy of object token selection.

Finally, future work should consider a more deterministic way to classify the concreteness of a token with NLP, including selection of definitions based on contextual awareness.

\section{Conclusion}
\label{sec:conclusion}
The continued challenge of visual grounding and masking in MMT systems has encouraged varying approaches to the detection and selection of visually-grounded text tokens for masking \cite{caglayan-etal-2019-probing, wu-good-2021}.

We introduced three new techniques for detection of concrete tokens from source sentences: detection with natural language processing (NLP), detection with object detection, and joint NLP/object detection. We also introduced deterministic methods for the selection of detected tokens, including longest and shortest $n$ tokens.

Finally, we utilized the GRAM MMT architecture \cite{vijayan-multimodal-2024} to train models against synthetically collated datasets of masked sentences and associated images. We showed performance improvement over the baseline models and elevated usage of visual context during translation tasks.

\thanksnostar{Opinions, interpretations, conclusions, and recommendations are those of the authors and are not necessarily endorsed by the United States Government. Cleared for public release on 12 Feb 2024. Originator reference number RH-24-125351. Case number AFRL-2024-0803.}

\newpage
\bibliography{bib/anthology,bib/sources}

\newpage
\appendix
\Large{\textbf{Appendix}}

\begin{table}[hbp]
    \centering
    \includegraphics[width=\linewidth, trim={2.5cm 4.5cm 2.5cm 0.75cm}, clip=true]{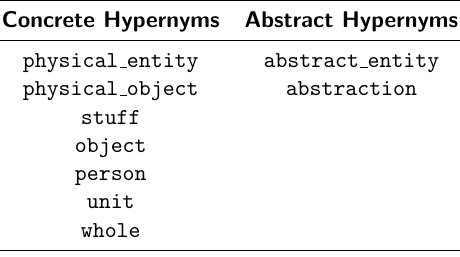}
    \caption{Labeled WordNet \cite{miller_wordnet_1995} hypernyms. A token is classified as concrete or abstract if any of the above hypernyms are in its DAG.}
    \label{tab:hyper}
\end{table}

\end{document}